\documentclass[runningheads]{llncs}

\usepackage{eccv}

\usepackage{eccvabbrv}

\usepackage{graphicx}
\usepackage{booktabs}
\usepackage{pifont}
\usepackage{array}
\usepackage{multirow}

\usepackage[accsupp]{axessibility}  

\usepackage{hyperref}

\usepackage{orcidlink}

\begin{document}

\title{End-To-End Underwater Video Enhancement: Dataset and Model} 

\titlerunning{End-To-End Underwater Video Enhancement}

\author{Dazhao Du\inst{1} \and
Enhan Li\inst{1} \and
Lingyu Si\inst{2,3} \and
Fanjiang Xu\inst{2,3} \and
Jianwei Niu\inst{1}
}

\authorrunning{Du et al.}

\institute{Hangzhou Innovation Institute, Beihang University \\
\email{\{dudazhao,lienhan,niujianwei\}@buaa.edu.cn}
\and
Institute of Software, Chinese Academy of Science \\
\email{\{lingyu,fanjiang\}@iscas.ac.cn} \and
University of Chinese Academy of Sciences
}

\maketitle

\begin{abstract}
Underwater video enhancement (UVE) aims to improve the visibility and frame quality of underwater videos, which has significant implications for marine research and exploration. However, existing methods primarily focus on developing image enhancement algorithms to enhance each frame independently. There is a lack of supervised datasets and models specifically tailored for UVE tasks. To fill this gap, we construct the \textbf{S}ynthetic \textbf{U}nderwater \textbf{V}ideo \textbf{E}nhancement (SUVE) dataset, comprising 840 diverse underwater-style videos paired with ground-truth reference videos. Based on this dataset, we train a novel underwater video enhancement model, UVENet, which utilizes inter-frame relationships to achieve better enhancement performance. Through extensive experiments on both synthetic and real underwater videos, we demonstrate the effectiveness of our approach. This study represents the first comprehensive exploration of UVE to our knowledge. The code is available at \url{https://anonymous.4open.science/r/UVENet}.
  \keywords{Underwater Video Enhancement \and Underwater Image Enhancement \and CNN}
\end{abstract}

\section{Introduction}
\label{sec:intro}

With the rise of marine ecological research, underwater robots, and underwater archaeology, the processing and understanding of underwater images and videos have attracted more attention~\cite{seathru,funiegan}. However, due to the impact of wavelength-dependent light absorption and scattering in water, underwater images and videos usually suffer from various visual degradations, such as color casts, blurred details, and low contrast~\cite{fusion,ghani2015enhancement}. Underwater image and video enhancement aims to obtain higher-quality and clearer images and videos from these degraded underwater data. 

\begin{figure}[tb]
  \centering
  \includegraphics[width=0.8\columnwidth]{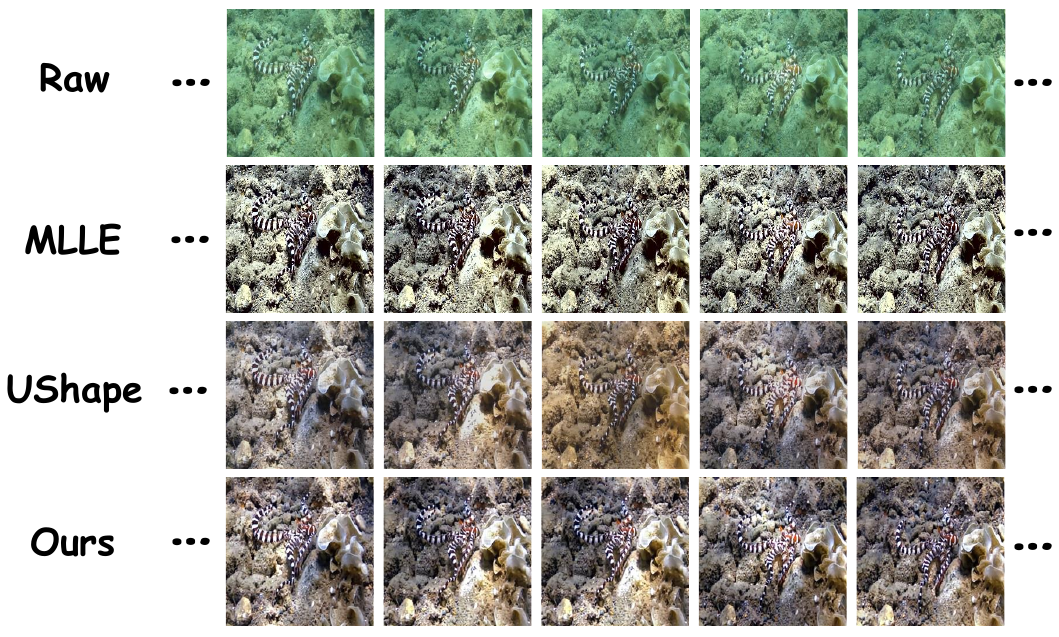}
  \caption{Visual results of different enhancement methods on a real underwater video. Only five frames in each video are shown for ease of display. From top to bottom, the four rows display the frames from the raw underwater video, the frame-by-frame enhancement results of MLLE, the frame-by-frame enhancement results of UShape, and our video enhancement results, respectively.}
  \label{fig:motivation}
\end{figure}

Recently, researchers have proposed many advanced underwater image enhancement (UIE) methods~\cite{fusion,mlle,uwcnn,ugan,utrans,uiec,uranker,fivenet} that achieved promising performance. However, there is little exploration of underwater video enhancement (UVE)~\cite{fusion,li2015simultaneous,qing2016underwater}. A simple solution is to directly extend these UIE methods to videos, independently enhancing each frame to obtain enhanced videos~\cite{uwcnn}. However, this solution does not exploit the temporal relationships among frames and can produce temporally inconsistent videos suffering from temporal artifacts and flickering. To illustrate this point, we apply two advanced UIE methods to enhance a real underwater video frame by frame. MLLE~\cite{mlle} directly adjusts the image colors, while UShape~\cite{utrans} is a Transformer-based enhancement model. As shown in \cref{fig:motivation}, the frames enhanced by MLLE exhibit poor quality, while the video generated by UShape displays noticeable flickering in the third frame. Other methods utilized temporal filtering~\cite{fusion} or spatial-temporal information fusion~\cite{qing2016underwater} to post-process the enhancement results of all the frames, but they cannot fundamentally solve the problem and cannot be trained end-to-end. To address these issues, we construct the first dataset specifically for UVE and design a novel model that takes multiple frames as input simultaneously and models the temporal relationships among frames.

Different from other low-level vision tasks such as super-resolution~\cite{chan2021basicvsr} and deblurring~\cite{su2017deep}, which only require downsampling and filtering of the original videos to construct paired videos, it is difficult or even impractical to simultaneously obtain real underwater videos and corresponding ground-truth reference videos of the same scene~\cite{waternet}. In the context of UIE, researchers typically employ two approaches to construct paired image datasets. The first involves applying multiple UIE algorithms to enhance each underwater image and manually selecting the clearest enhanced result as the ground-truth reference image~\cite{waternet,utrans}. However, applying this approach frame-by-frame to underwater videos can also result in reference videos with temporal artifacts. Therefore, we explore an alternative approach: simulating the underwater imaging process to synthesize underwater-style videos based on in-air videos~\cite{watergan,uwgan,uwcnn,UWNR}. This approach ensures temporal consistency in the synthesized underwater video by controlling the consistency of imaging parameters across all frames. Specifically, we utilize underwater neural rendering (UWNR)~\cite{UWNR} and control the consistency of the light field maps to synthesize underwater-style video for any high-quality indoor video in the NYU-Depth V2 dataset~\cite{nyudepth}. Through this approach, we construct the first synthetic underwater video enhancement dataset (SUVE) which covers diverse water types and lighting conditions. SUVE includes 840 pairs of videos and each video pair consists of an underwater-style video and its corresponding reference video.

With the constructed SUVE dataset, we can train a UVE model end-to-end. Inspired by general video restoration tasks~\cite{su2017deep,chan2021basicvsr,li2023simple,xu2023video}, we design a novel underwater video enhancement model, UVENet, which can handle multi-frame inputs simultaneously. UVENet utilizes a shared convolutional backbone~\cite{liu2022convnet} to extract multi-scale features from all input frames. Due to the strong spatial relationships among adjacent frames, we introduce core Feature Alignment and Aggregation Modules (FAAMs), which employ grouped spatial shifts~\cite{li2023simple} to align multi-frame features and aggregate them using depth-wise separable convolution. Finally, the enhanced frames are generated based on the aggregated features through a decoder and a Global Restoration Module (GRM).

Our contributions can be summarized as follows:
\begin{enumerate}
    \item We construct the first synthetic underwater video enhancement (SUVE) dataset, which includes 840 pairs of underwater-style videos and their corresponding ground-truth videos for training and evaluating UVE models.
    \item We design a novel model, UVENet, which utilizes FAAMs to capture inter-frame relationships and a GRM to enhance visual effects. The model can be trained end-to-end on the SUVE dataset.
    \item Extensive experiments on both synthetic and real conditions demonstrate that our method outperforms the state-of-the-art UIE methods in terms of enhanced frame quality and inter-frame continuity.
\end{enumerate}

\section{Related Work}

\subsubsection{Underwater Image Enhancement.}
Existing UIE methods can be divided into three categories: physical model-based, physical model-free, and data-driven methods~\cite{du2023uiedp}. Physical model-based methods estimate the parameters of the underwater image formation model to invert the degradation process~\cite{udcp,seathru,galdran2015automatic}. Physical model-free methods directly adjust image pixel values by histogram equalization~\cite{histogram}, color balance~\cite{mlle,fusionv2}, and contrast correlation~\cite{ghani2015enhancement}. Recently, researchers have constructed paired image datasets for UIE by synthesizing ground truth images~\cite{waternet,utrans} or underwater images~\cite{watergan,uwgan,uwcnn,UWNR}. The introduction of these datasets has led to a dominance of data-driven methods. Various advanced deep models~\cite{ucolor,uiec,fivenet,utrans,pugan,waternet} tailored to underwater scenes have achieved significant enhancement performance. For example, Li et al.~\cite{ucolor} presented an underwater image enhancement network via medium transmission-guided multi-color space embedding. Peng et al.~\cite{utrans} used Transformers to fusion channel-wise multi-scale features and model spatial-wise global features. FA$^+$Net~\cite{fivenet} designed color enhancement modules and pixel attention modules to amplify the network’s perception of details. However, these methods mentioned above only consider the spatial characteristics within a single image but not the temporal characteristics between the video frames, thus making them unsuitable for directly enhancing underwater videos.

\subsubsection{Underwater Video Enhancement.}
Currently, most methods for enhancing underwater videos are extensions of single image enhancement algorithms~\cite{hu2022overview}, which enhance the video in a frame-by-frame manner. These methods focus on improving efficiency by designing fast filtering algorithms~\cite{lu2013underwater,tang2019efficient} or lightweight deep network models~\cite{uwcnn,funiegan}. However, independently processing each frame can lead to temporal artifacts and interframe flicker phenomena. Some methods consider the temporal relationships among frames. Ancuti et al.~\cite{fusion} used the time-bilateral filtering strategy to achieve smoothing between frames and maintain temporal coherence. Li et al.~\cite{li2015simultaneous} proposed a new video dehazing algorithm and extended it to remove the hazy phenomena in underwater videos. To dehaze the underwater videos, Qing et al.~\cite{qing2016underwater} estimated consistent transmission and background light by a spatial–temporal information fusion method. However, these methods cannot be trained end-to-end. Some recent studies~\cite{zhou2023cross,varghese2023self} have utilized adjacent frames as additional views to assist in UIE, yet they are not specifically designed for UVE. In this paper, we construct the first paired video dataset for UVE and propose a new video enhancement model.

\subsubsection{General Video Restoration.}
Other video restoration tasks, such as video super-resolution~\cite{chan2021basicvsr}, deblurring~\cite{su2017deep}, denoising~\cite{li2023simple}, dehazing~\cite{xu2023video}, and low-light enhancement~\cite{MABD}, have made notable progress. These advancements can be attributed to two factors: large-scale paired video datasets~\cite{su2017deep,nah2019ntire,xu2023video} available for supervised training and advanced video restoration models~\cite{chan2021basicvsr,li2023simple}. Unlike these tasks, the complicated underwater imaging environment makes it difficult to obtain paired underwater video datasets. We overcame this difficulty by synthesizing realistic underwater videos. As for video restoration models, the key lies in the design of various alignment modules to capture inter-frame relationships and aggregate temporal information, such as optical flow estimation~\cite{xue2019video,chan2021basicvsr}, deformable convolutions~\cite{wang2019edvr,tian2020tdan}, and cross-frame self-attention layers~\cite{liang2022vrt}. It is worth noting that a recent work, Shift-Net~\cite{li2023simple}, introduced a grouped spatial-temporal shift to implicitly capture inter-frame correspondences for multi-frame aggregation. Inspired by these approaches, we propose the first video enhancement model tailored for underwater scenes.

\section{SUVE Dataset}
\label{sec:suve}

\begin{figure}[tb]
  \centering
  \includegraphics[height=6.5cm]{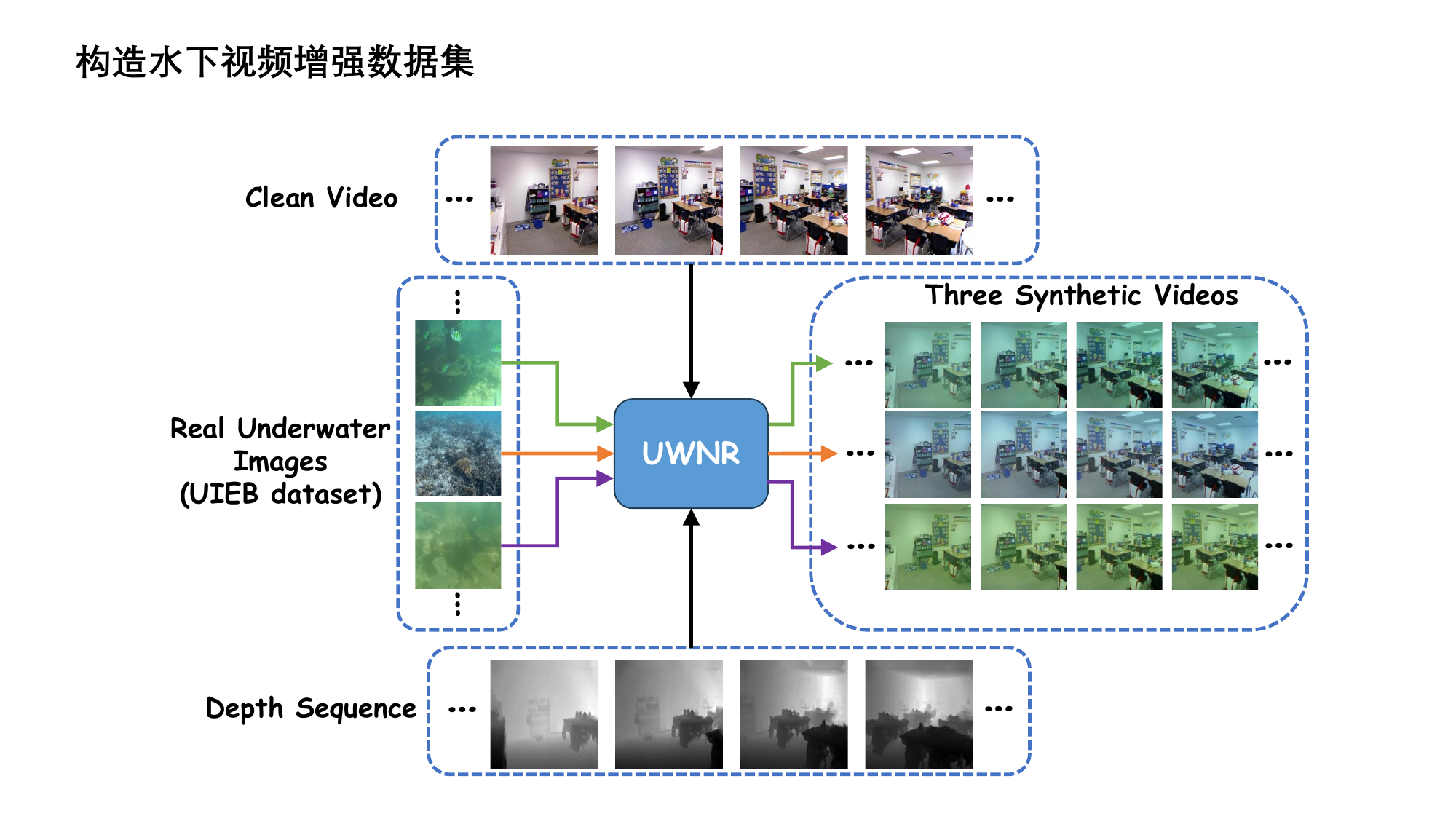}
  \caption{The synthesis process of the SUVE dataset. Given a clean in-air video, the corresponding depth sequence, and a randomly sampled real underwater image, UWNR can synthesize underwater video with a similar style to the sampled underwater image.}
  \label{fig:suve}
\end{figure}

Since the current paired underwater datasets are mostly collected for underwater image enhancement (UIE), we construct the first synthesized underwater video enhancement (UVE) dataset named SUVE.

Researchers typically employ two approaches to generate paired datasets in UIE. One approach involves employing multiple UIE algorithms to enhance each real underwater image and manually selecting the visually optimal result as the reference image (also known as ground truth)~\cite{waternet,utrans}. However, different frames from the same video may correspond to different optimal enhancement algorithms. If the optimal result is selected for each frame in this manner, the resulting ground truth video will suffer from artifacts and flickers. The alternative approach utilizes clean images, corresponding depth maps, and a degradation model to generate underwater-style images. For example, some use underwater imaging physical models~\cite{uwcnn}, GANs~\cite{watergan,uwgan}, or implicit degradation model~\cite{UWNR} to synthesize underwater images. Since the underwater style can be controlled by the parameters of the degradation model, we can generate underwater images for all frames of each video by applying the same parameters, which allows us to create temporally consistent underwater videos. We introduce the specific procedure for constructing the SUVE dataset below.

The original NYU-Depth V2 dataset~\cite{nyudepth} consists of 280 high-quality video sequences and corresponding depth maps collected from various indoor scenes. We use these clean videos as the ground truth to synthesize the corresponding underwater videos. As the preprocessing step, we applied cross-bilateral filters at different scales to fill the missing depth values in raw depth maps. For all video frames sized 480$\times$640, we performed center cropping to 460$\times$620 to remove white borders from the images. We divided the 280 videos into a training set of 220 videos and a test set of 60 videos. Next, we synthesize the corresponding underwater videos for each clean video in the training and test sets.

Considering the quality and controllability, we generate underwater videos following UWNR (Underwater Neural Rendering)~\cite{UWNR}. Given a clean image, its corresponding depth map, and a light field map containing expected underwater scene characteristics, the pre-trained UWNR model can output a high-quality underwater-style image. All frames from the same video share the same light field map derived from one real underwater image to generate one temporally consistent underwater video. For each clean video, we repeat the above process three times by randomly selecting three real underwater images (three light field maps) from the UIEB dataset~\cite{waternet}. Therefore, each ground truth video corresponds to three synthesized underwater videos with different underwater styles. Overall, we obtain 660 and 180 pairs of synthetic videos in training and testing sets, respectively. \cref{fig:suve} shows the process of generating underwater videos and examples of paired videos. More examples can be found in the appendix.

\section{Methodology}
\subsection{Overview of UVENet}
\begin{figure}[tb]
  \centering
  \includegraphics[width=0.9\columnwidth]{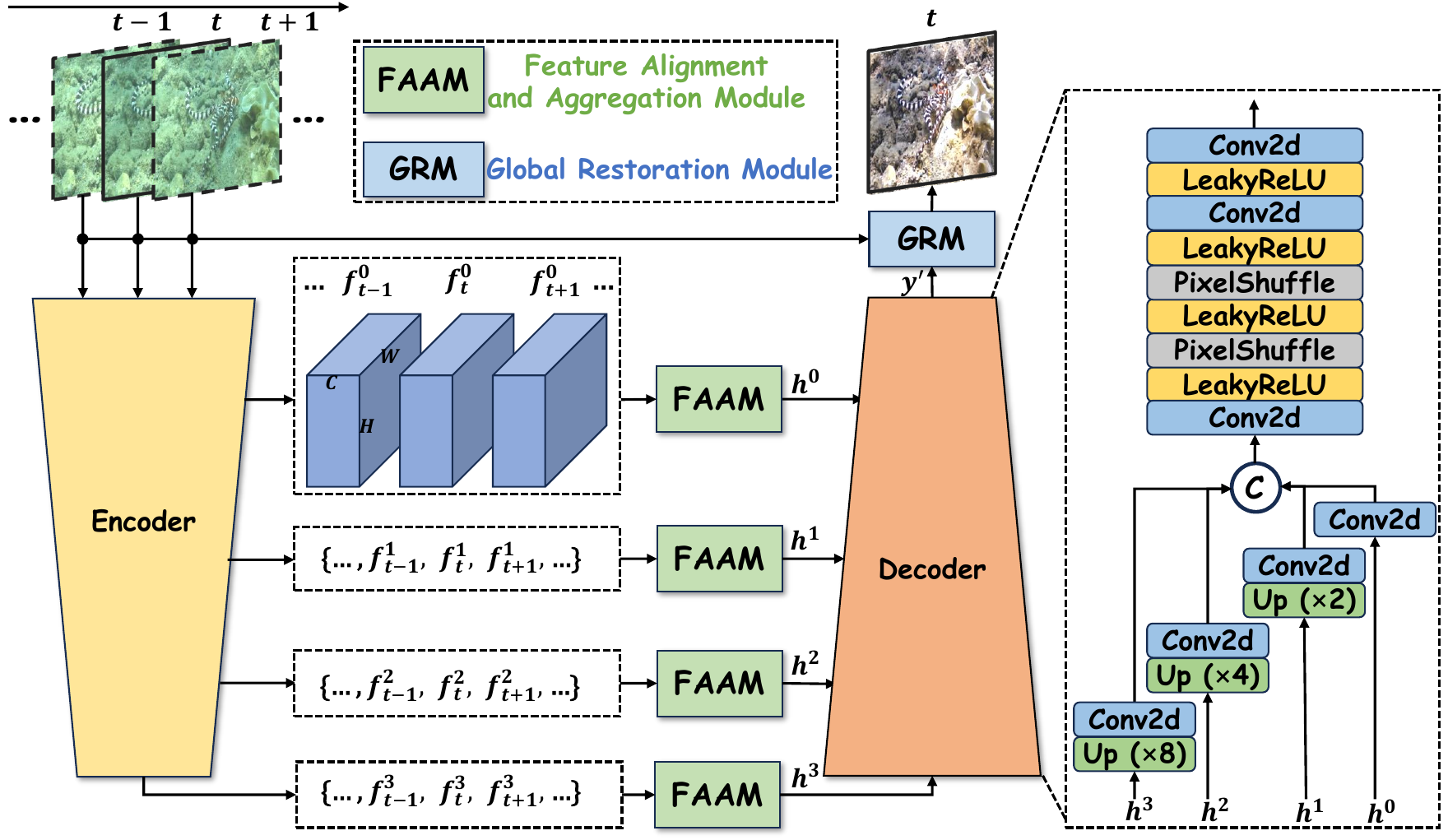}
  \caption{The overall framework of our UVENet. It mainly consists of an encoder, four feature alignment and aggregation modules dealing with feature maps at different scales, a decoder, and a global restoration module. For ease of display, we only plot the case where there are three input frames, i.e., $T=3$.}
  \label{fig:uvenet}
\end{figure}

Given consecutive frames $\{x_i\in \mathbb{R}^{3 \times H\times W}\}_{i=t-k}^{i=t+k}$ in an underwater video, where $t$ is the index of the center frame and $T=2k+1$ denotes the number of all input frames, UVENet aims to output the high-quality version of the center frame, $y_t$. As shown in \cref{fig:uvenet}, UVENet mainly consists of four parts: an encoder, four feature alignment and aggregation modules (FAAMs), a lightweight decoder, and a global restoration module. We describe them in detail in the following subsections.

\subsection{Encoder}
The encoder is responsible for extracting frame-wise features from the input frames, which can be implemented by any powerful image backbone. Here we employ ConvNeXt~\cite{liu2022convnet} to extract the multi-scale frame-wise feature maps. We find that Instance Normalization~\cite{instancenorm} is more suitable for enhancement tasks compared to Layer Normalization~\cite{layernorm}. Therefore, we replace all the Layer Normalization in ConvNeXt with Instance Normalization. Note that the encoder is shared for all frames. For all input frames $\{x_i\}_{i=t-k}^{i=t+k}$, the encoder outputs feature maps of four different scales, $\{f_i^s\}_{i=t-k}^{i=t+k}$, where $s=0,1,2,3$ is the index of scale. As the scale $s$ increases, the spatial resolution of the feature maps halves, while the number of channels doubles. Subsequently, the feature map sequences at four scales are fed into four independent FAAMs for further processing.

\subsection{Feature Alignment and Aggregation Module}

\begin{figure}[tb]
  \centering
  \includegraphics[width=0.85\columnwidth]{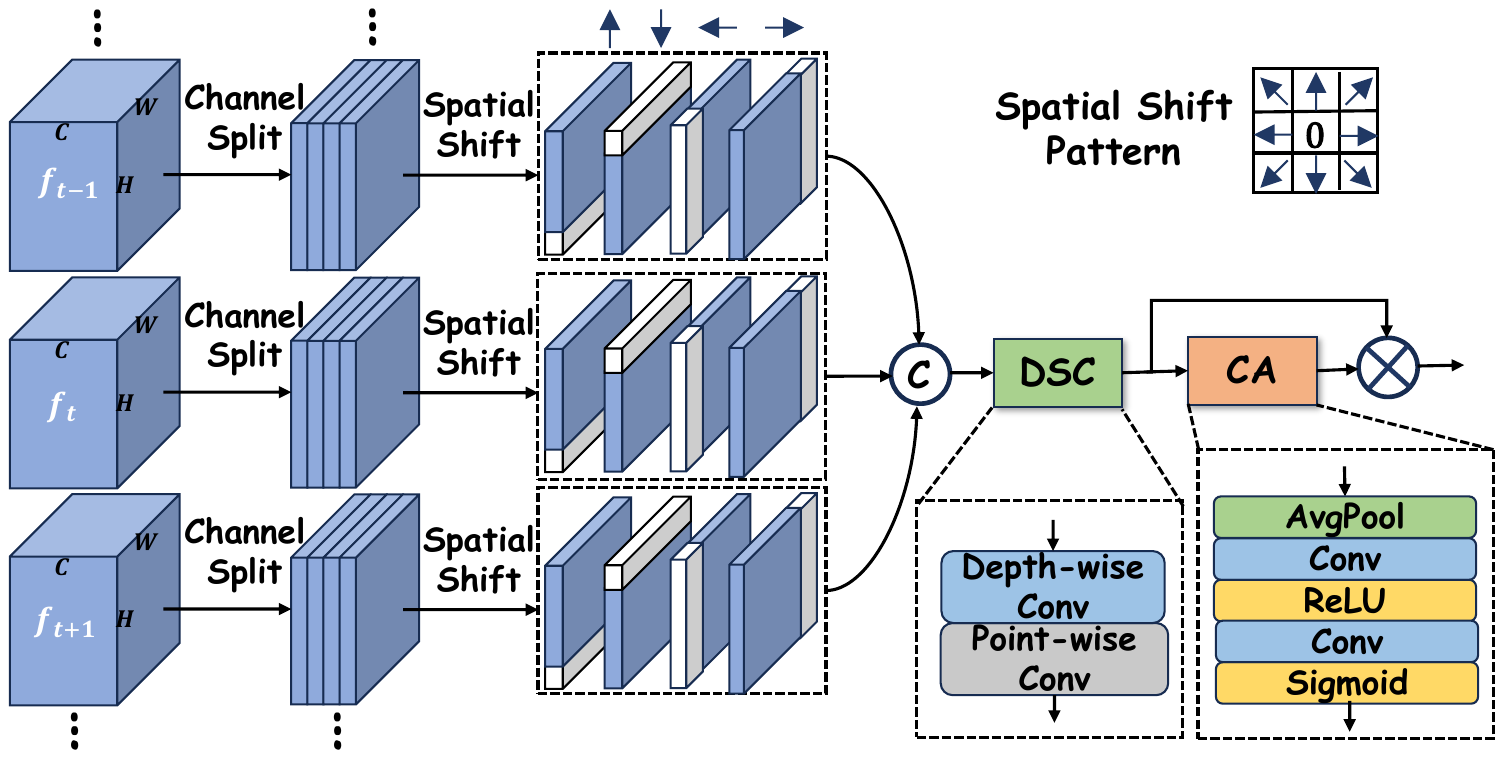}
  \caption{The illustration of the Feature Alignment and Aggregation Module (FAAM). We only draw the feature maps of three adjacent frames and four spatial shift patterns (up, down, left, right). In practice, we have four additional patterns for diagonal shifts. The void pixels (white parts in the shifted feature maps) caused by spatial shifts are filled with zeros. DSC and CA denote Depth-wise Separable Convolution and Channel Attention, respectively.}
  \label{fig:faam}
\end{figure}

Due to spatial offsets among frames in a video, it is crucial to perform spatial alignment among features of adjacent multiple frames, which helps in learning the dynamic evolution, removing inter-frame redundancy, and supplementing inter-frame information. We do not use common optical flow~\cite{xue2019video,chan2021basicvsr} or deformable convolutions~\cite{wang2019edvr,tian2020tdan} for feature alignment, but instead adopt a simpler and more efficient grouped spatial shift operation~\cite{li2023simple}. As shown in~\cref{fig:faam}, we first equally split (i.e. group) the $i$-th frame feature $f_i \in \mathbb R ^{C \times H\times W}$ along the channel dimension to obtain eight feature slices $f_{i,m} \in \mathbb R ^{\frac{C}{8} \times H\times W}$, where $m=1,...,8$.%
\footnote{For the sake of simplicity, we have omitted the superscript representing the scale index and only plotted four feature slices.}
Subsequently, each feature slice $f_{i,m}$ is spatially shifted by $(l*\Delta x_m, l*\Delta y_m)$ pixels in the $x$ and y directions to obtain the shifted feature slice $f'_{i,m}$, where $l$ is the base length of spatial shift and $\Delta x_m, \Delta y_m \in \{-1,0,1\}$. For example, when $\Delta x_m=0, \Delta y_m=1$, the feature slice is spatially shifted downward by $l$ pixels. Therefore, there are a total of eight spatial shift patterns corresponding to eight alignment direction candidates (excluding the case of no shift), as shown in~\cref{fig:faam}. After shifting, we concatenate all feature slices along the channel dimension to obtain the spatially shifted $i$-th frame feature $f'_{i}$. 

To fuse information across multiple frames, we utilize Depth-wise Separable Convolution (DSC)~\cite{howard2017mobilenets} and Channel Attention (CA)~\cite{hu2018squeeze} to aggregate shifted multi-frame features $\{f'_{i}\}_{i=t-k}^{i=t+k}$. Specifically, these features are first concatenated along the channel dimension and then fed into a parameter-efficient DSC to obtain the preliminary aggregated feature. Subsequently, the preliminary aggregated feature is refined by CA to emphasize the channel slices corresponding to actual spatial shifts, resulting in the final aligned and aggregated feature $h$. 

As depicted in~\cref{fig:uvenet}, we apply the above feature alignment and aggregation module (FAAM) on feature sequences at all four scales to cover sufficient spatial shifts. Finally, we derive four aligned and aggregated features of different scales, denoted as $\{h^0,h^1,h^2,h^3\}$, with the superscript indicating the scale index.

\subsection{Decoder}
We utilize a lightweight decoder to reconstruct the high-quality image from the multi-scale feature maps $\{h^0,h^1,h^2,h^3\}$ processed by FAAMs. The specific structure of the decoder is illustrated on the right side of~\cref{fig:uvenet}. Different upsampling factors are applied to feature maps at different scales to bring them to the same spatial resolution. We employ bilinear interpolation for upsampling followed by an additional convolutional layer. The four upsampled feature maps are concatenated along the channel dimension and fed into a compact network comprising convolutions, non-linear activation functions, and pixel shuffle layers~\cite{shi2016real} to generate the preliminary enhanced image $y'$ with the same spatial resolution as the input frames.

\subsection{Global Restoration Module}

\begin{figure}[tb]
  \centering
  \includegraphics[width=0.85\columnwidth]{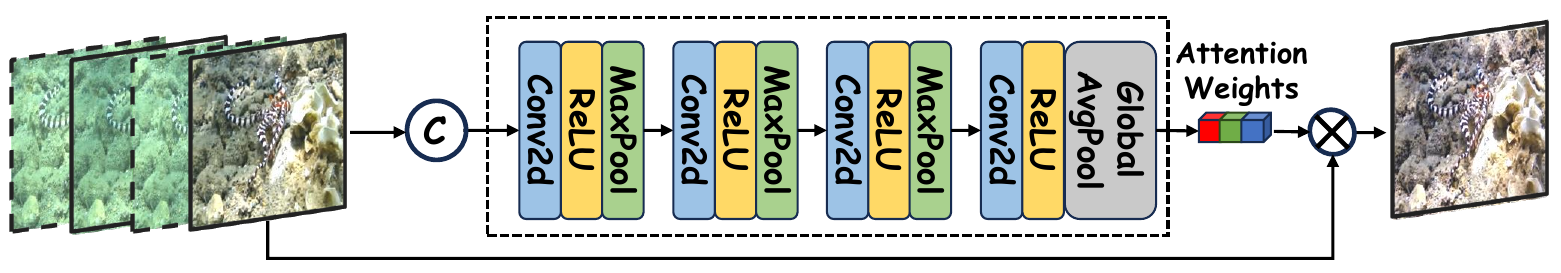}
  \caption{The illustration of the Global Restoration Module (GRM). Low-quality frames and the preliminary enhanced image are concatenated along the channel dimension and fed into the attention module to generate attention weights. The attention weights correspond to scaling factors for the RGB three channels, which are applied to the preliminary enhanced image to obtain the final enhanced frame image.}
  \label{fig:grm}
\end{figure}

Inspired by the Gray World algorithm, which has been effectively applied in UIE~\cite{fu2017two,mlle}, we propose a global restoration module (GRM) for color adjustment to further improve the visual quality of enhanced frames. The structure of GRM is shown in~\cref{fig:grm}, where the coefficients for color adjustment in the Gray World algorithm are replaced by adaptive attention weights. Firstly, we concatenate the preliminary enhanced image $y'$ and the low-quality underwater frames along the channel dimension and feed them into an attention module composed of multiple convolutional layers, non-linear activation functions, and pooling layers. The last pooling layer is a global average pooling layer, which is used to generate global attention weights corresponding to the RGB three channels. Finally, the attention weights are element-wise multiplied with $y'$ by the broadcasting mechanism to obtain the final enhanced image $y_t$.

\section{Experiments}

\subsection{Datasets}
\subsubsection{Synthetic Dataset.} We synthesized the paired underwater video enhancement dataset SUVE according to~\cref{sec:suve}. SUVE comprises 660 pairs of training videos and 180 pairs of testing videos, covering a wide range of underwater imaging conditions. Each video contains an average of around 170 frames, resulting in over 140,000 underwater images in the entire dataset.
\subsubsection{Real Dataset.} Original MVK (Marine Video Kit) dataset~\cite{mvk} is composed of many real underwater videos from different locations worldwide and at different times across the year. To evaluate this dataset, we selected 45 challenging videos with durations ranging from 7 seconds to 3 minutes, covering 36 different locations. The average length of these 45 videos is around 30 seconds. In subsequent experiments, we downsampled the videos to 5 frames per second. As a result, the entire dataset comprises over 7000 frame images.
\subsection{Evaluation Metrics}
\subsubsection{Frame-level Metrics.} We evaluate the quality of each frame image in enhanced videos and report the average metrics of all frames. PSNR and SSIM are commonly used full-reference image quality evaluation metrics, which measure the similarity between enhanced images and reference images. Besides, we have adopted two non-reference image quality assessment metrics specifically designed for underwater images: UIQM~\cite{UIQM} and UCIQE~\cite{UCIQE}. UIQM considers contrast, sharpness, and colorfulness of images, while UCIQE considers brightness instead of sharpness.
\subsubsection{Video-level Metrics.} We evaluate the temporal consistency and artifacts in videos from the perspectives of brightness and color. The mean absolute brightness differences (MABD) vector reflects the general level of time derivatives of brightness values at each pixel location. In the low-light video enhancement task, the mean square error (MSE) between MABD vectors of an enhanced video and that of the ground truth video could serve as an indication of its flickering effect~\cite{MABD}. In the video colorization task, the Color Distribution Consistency index (CDC)~\cite{CDC} is employed to measure the temporal consistency, which computes the Jensen-Shannon divergence of the color distribution between consecutive frames. In this study, we directly apply these two metrics to assess the quality of enhanced underwater videos.

\begin{figure}[t]
  \centering
  \includegraphics[width=0.99\columnwidth]{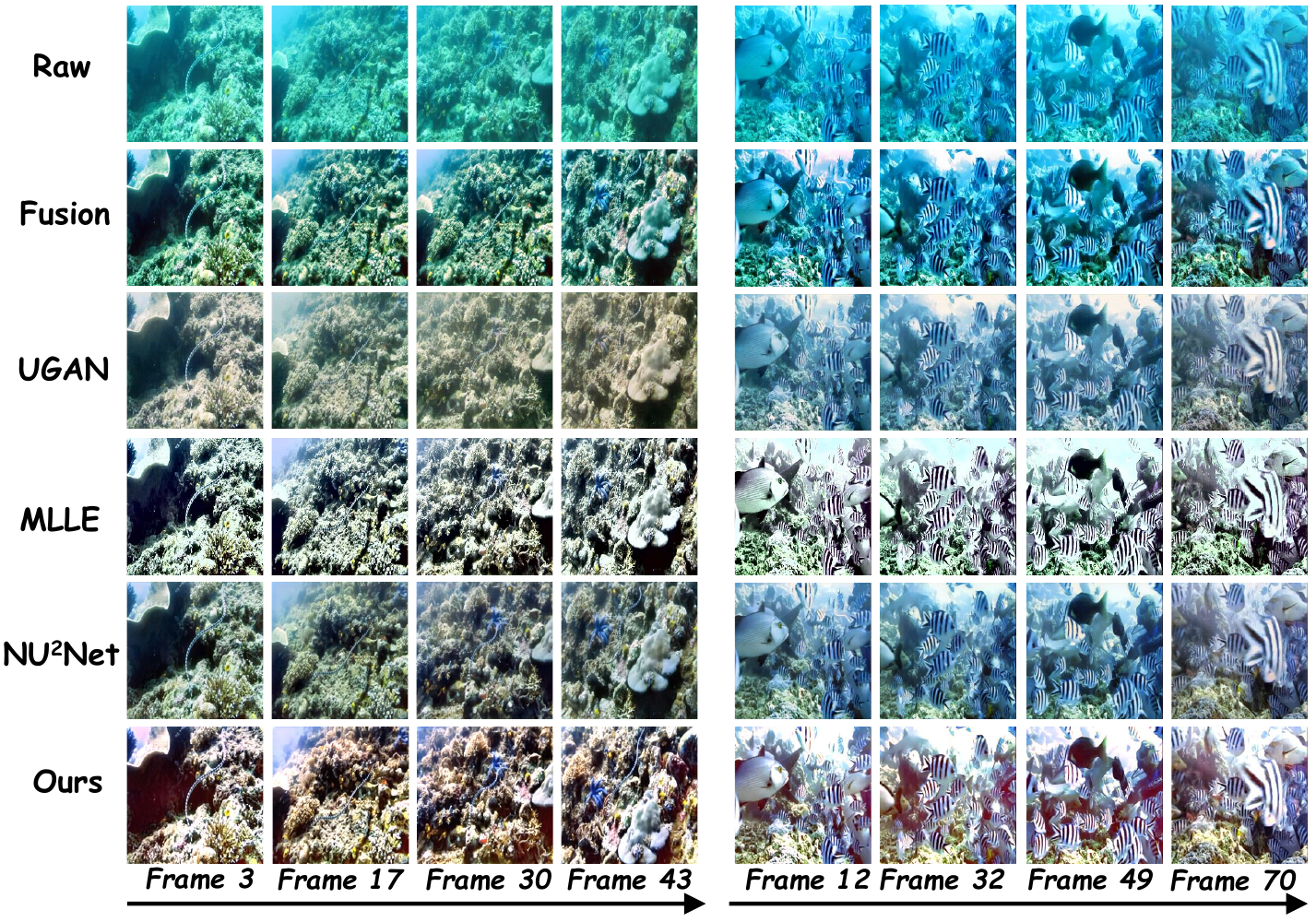}
  \caption{Visual comparison of our UVENet with several UIE methods on two real underwater videos from the MVK dataset. We randomly select four frames in each video for ease of display.}
  \label{fig:exp_mvk}
\end{figure}

\subsection{Implementation Details}
For our UVENet, we collect training samples by randomly sampling windows with a size equal to the number of input frames $T=5$. We adopt Adam optimizer~\cite{adam} and Cosine Annealing scheme. The initial learning rate is set to $4\times 10^{-4}$. The total number of iterations is 80K and the batch size is 16. All the frames are resized to 256$\times$256 during both training and testing. Besides, data augmentation techniques such as horizontal flipping and rotation are employed during training. For inference, we employ a sliding window approach following~\cite{wang2019edvr}. The base shift length $l$ in FAAMs is set to 3. More details about the UVENet model can be found in the appendix. We replicate all the compared UIE methods based on their official codes and hyperparameters. For these UIE models, we train them using all frames from the SUVE training set for 50 epochs, which is sufficient for any UIE model to converge. We use L1 loss to train all models.

\subsection{Performance Evaluation}

\begin{table}[t]
\caption{Comparison with the state-of-the-art UIE methods on our SUVE dataset. The top two results in each column are marked in \textcolor{red}{red} and \textcolor{blue}{blue}, respectively.}
\label{tab:sota_suve}
\centering
\begin{tabular}{l|cccccc}
\toprule
\textbf{Methods} & PSNR$\uparrow$  & SSIM$\uparrow$ &  UCIQE$\uparrow$ & UIQM$\uparrow$ & MSE(MABD) $\downarrow$ & CDC$\downarrow$ \\
\midrule
Fusion~\cite{fusion}  & 17.06 & 0.783 & \textcolor{red}{0.607} & 2.800 & 25.09 & \textcolor{blue}{0.0114}\\
UWCNN~\cite{uwcnn}   & 22.51 & 0.862 & 0.549 & 2.745 & 11.68 & 0.0268\\
UGAN~\cite{ugan}    & 20.02 & 0.811 & 0.577 & \textcolor{blue}{3.108} & 16.42 & 0.0123\\
FUnIEGAN~\cite{funiegan}  & 19.15 & 0.779 & 0.581 & \textcolor{red}{3.164} & 20.31 & 0.0136\\
MLLE~\cite{mlle}      & 16.88 & 0.675 & 0.591 & 1.996 & 178.74 & \textcolor{red}{0.0084}\\
UShape~\cite{utrans} & 24.37 & 0.878 & 0.574 & 2.698 & 8.08 & 0.0367\\
UIEC$^2$Net~\cite{uiec}  & 25.72 & 0.899 & 0.581 & 2.723 & \textcolor{blue}{0.99} & 0.0129\\
NU$^2$Net~\cite{uranker} & \textcolor{blue}{26.57} & \textcolor{blue}{0.906} & 0.575 & 2.747 & 1.67 & 0.0145\\
FA$^{+}$Net~\cite{fivenet} & 23.90 & 0.874 & 0.569 & 2.773 & 5.54 & 0.0186\\
\midrule
\textbf{UVENet}  & \textcolor{red}{27.54} & \textcolor{red}{0.919} & \textcolor{blue}{0.592} & 2.595 & \textcolor{red}{0.86} & 0.0127\\
\bottomrule
\end{tabular}
\end{table}

\begin{table}[t]
\caption{Comparison with the state-of-the-art UIE methods on the MVK dataset. The top two results are marked in \textcolor{red}{red} and \textcolor{blue}{blue}, respectively.}
\label{tab:sota_mvk}
\centering
\begin{tabular}{l|ccc||l|ccc}
\toprule
\textbf{Methods} & UCIQE$\uparrow$ & UIQM$\uparrow$   &  CDC$\downarrow$ \hspace{0.2cm}&  \textbf{Methods} & UCIQE$\uparrow$ & UIQM$\uparrow$   &  CDC$\downarrow$  \\
\midrule
Fusion~\cite{fusion} & \textcolor{red}{0.634} & 2.788 & 0.0146 \hspace{0.2cm}& UShape~\cite{utrans} & 0.559 & 2.853 & 0.0299  \\
UWCNN~\cite{uwcnn} & 0.515 & 2.760 & 0.0375 \hspace{0.2cm}&  UIEC$^2$Net~\cite{uiec} & 0.582 & \textcolor{blue}{3.012} & 0.0201\\
UGAN~\cite{ugan} &  0.558 & \textcolor{red}{3.030} & 0.0159 \hspace{0.2cm}& NU$^2$Net~\cite{uranker} & 0.561 & 2.981 & 0.0204 \\
FUnIEGAN~\cite{funiegan} &  0.555 & 2.974 & 0.0170 \hspace{0.2cm}& FA$^{+}$Net~\cite{fivenet} & 0.529 & 2.840 & 0.0324 \\
MLLE~\cite{mlle} &  0.600 & 2.100 & \textcolor{blue}{0.0125} \hspace{0.2cm}&  \textbf{UVENet} & \textcolor{blue}{0.611} & 2.806  & \textcolor{red}{0.0118}  \\
\bottomrule
\end{tabular}
\end{table}

We compare our method with several state-of-the-art UIE methods, including two physical model-free methods (Fusion~\cite{fusion}, MLLE~\cite{mlle}) and eight data-driven methods (UWCNN~\cite{uwcnn}, UGAN~\cite{ugan}, FUnIEGAN~\cite{funiegan}, UShape~\cite{utrans}, FA$^+$Net~\cite{fivenet}, UIEC$^2$Net~\cite{uiec}, NU$^2$Net~\cite{uranker}). These UIE methods enhance underwater videos frame by frame. Results for two datasets are presented in~\cref{tab:sota_suve,tab:sota_mvk}. For an intuitive comparison, \cref{fig:exp_mvk} visualizes the enhanced results of several methods on the MVK dataset. More visual results are available in the supplementary material.

For frame-level metrics, our method outperforms all other UIE methods in both PSNR and SSIM. Fusion performs the best on the UCIQE metric, while UGAN excels on the UIQM metric. However, from a human visual perspective, the enhancement results of Fusion and UGAN shown in~\cref{fig:exp_mvk} are not satisfactory. This observation indicates that these two non-reference quality assessment metrics designed for underwater images are inconsistent with human visual perception in some cases. The above observation has also been pointed out in previous works~\cite{waternet,pugan}. Conversely, even though our model performs poorly on the UIQM metric, it provides the clearest and most visually pleasing results.

For video-level metrics, MLLE performs the best in the CDC metric on the SUVE dataset. However, as shown in~\cref{fig:exp_mvk}, MLLE overlooks the inherent color differences in the original video and tends to convert images into black-and-white tones, which makes the color differences among adjacent frames less noticeable, resulting in a lower CDC metric. Also, for this reason, MLLE performs poorly on the MSE(MABD) metric. Other UIE methods do not take into account the temporal continuity between frames, hence performing poorly on metrics measuring temporal consistency. For example, \cref{fig:exp_mvk} shows that there exist temporal artifacts and flickering in the enhancement results of UGAN and NU2Net on two videos. In contrast, our video enhancement method achieves the best MSE(MABD) metric on the SUVE dataset and the best CDC metric on the MVK dataset. 

Overall, our UVENet outperforms other UIE methods in both frame image quality and video temporal continuity, which aligns with our motivation and demonstrates the superiority of UVE over UIE.

\subsection{Ablation Studies}

\begin{table}[t]
  \caption{The impact of different numbers of input frames $T$ on the model performance. When $T=1$, the model degrades from video enhancement to image enhancement. GPU Mem. represents the GPU memory consumed during training when the batch size is 4.}
  \label{tab:abla_t}
  \centering
  \begin{tabular}{>{\centering\arraybackslash}p{1cm}|ccccccc}
    \toprule
    $T$ & PSNR$\uparrow$  & SSIM$\uparrow$ &  UCIQE$\uparrow$ & UIQM$\uparrow$ & MSE(MABD)$\downarrow$ & CDC$\downarrow$ &  GPU Mem. \\
    \midrule
    1 & 26.92 & 0.906 & 0.586 & 2.581 & 1.23 & 0.0140 & 7.8G\\
    3 & 27.18 & 0.916 & 0.586 & 2.588 & 1.02 & 0.0139 & 12.9G\\
    5 & 27.54 & 0.919 & \textbf{0.592}& \textbf{2.595} & 0.86 & 0.0127 & 18.2G\\
    7 & \textbf{27.56} & \textbf{0.920} & 0.589 & 2.569 & \textbf{0.73} & \textbf{0.0121} & 25.4G\\
  \bottomrule
  \end{tabular}
\end{table}

\subsubsection{Effect of the Number of Input Frames.} 
$T$ represents the number of input frames processed simultaneously by our UVENet. Theoretically, the more adjacent frames the model can perceive, the more information it can obtain, which naturally leads to superior enhancement effects. The results presented in~\cref{tab:abla_t} confirm this viewpoint. When $T=1$, the model essentially enhances the video frame by frame independently, resulting in the poorest performance. As $T$ increases, the model can capture more inter-frame relationships, leading to improvements in all metrics. This observation further emphasizes the significance of our research on UVE over UIE. However, increasing the number of input frames $T$ comes with higher training costs and increased GPU memory consumption. Therefore, we select $T=5$ as the default setting for the remaining experiments to achieve a trade-off between performance and efficiency.

\begin{table}[tb]
  \caption{The impact of applying FAAMs to feature maps at different scales ($f^0,f^1,f^2,f^3$ in~\cref{fig:uvenet}). When not using FAAM on any scale (the first row), the model does not capture the inter-frame relationships.}
  \label{tab:abla_faam}
  \centering
  \begin{tabular}{cccc|cccccc}
    \toprule
    $f^0$ & $f^1$ & $f^2$ & $f^3$ & PSNR$\uparrow$  & SSIM$\uparrow$ &  UCIQE$\uparrow$ & UIQM$\uparrow$ & MSE(MABD)$\downarrow$ & CDC$\downarrow$\\
    \midrule
    \ding{55} & \ding{55} & \ding{55} & \ding{55} & 26.81 & 0.908 & 0.584 & 2.553 & 1.29 & 0.0142\\
    \ding{51} & \ding{55} & \ding{55} & \ding{55} & 27.18 & 0.912 & 0.587 & 2.570 & 1.24 & 0.0136\\
    \ding{51} & \ding{51} & \ding{55} & \ding{55} & 27.44 & 0.918 & 0.585 & \textbf{2.599} & 0.93 & 0.0139\\
    \ding{51} & \ding{51} & \ding{51} & \ding{55} & 27.37 & \textbf{0.920} & 0.588 & 2.571 & \textbf{0.78} & 0.0135\\
    \ding{51} & \ding{51} & \ding{51} & \ding{51} & \textbf{27.54} & 0.919 & \textbf{0.592} & 2.595 & 0.86 & \textbf{0.0127}\\
  \bottomrule
  \end{tabular}
\end{table}
\begin{table}[tb]
  \caption{The impact of different base shift length $l$ and aggregation methods on the model performance in FAAM. \textit{Depth-wise} and \textit{Point-wise} indicate aggregation by only depth-wise convolution or point-wise convolution. \textit{DSC} represents depth-wise separable convolution, while \textit{DSC+CA} represents incorporating additional channel attention.}
  \label{tab:abla_faam1}
  \centering
  \begin{tabular}{>{\centering\arraybackslash}p{1cm}|c|cccccc}
    \toprule
    $l$   & Aggregation & PSNR$\uparrow$  & SSIM$\uparrow$ &  UCIQE$\uparrow$ & UIQM$\uparrow$ & MSE(MABD)$\downarrow$ & CDC$\downarrow$\\
    \midrule
    1     & \multirow{4}[2]{*}{DSC+CA} &   27.41  & 0.918 & 0.587 & 2.550 &0.91 & 0.0128 \\
    2     &  & 27.38 & 0.917 & 0.589 & \textbf{2.617} & 0.84 & 0.0132  \\
    3     &  & \textbf{27.54} & \textbf{0.919} & \textbf{0.592} & 2.595 & 0.86 & 0.0127 \\
    4     &  & 27.43 & 0.919 & 0.589 & 2.613 & \textbf{0.80} & \textbf{0.0126}  \\
    \midrule
    \multirow{4}[2]{*}{3} & Depth-wise & 27.12 & 0.911 & 0.585 & \textbf{2.621} & 1.24 & 0.0138 \\
          & Point-wise &  27.03 & 0.913 & 0.588 & 2.613 & 1.15 & 0.0133  \\
          & DSC   & 27.43 & 0.918 & 0.586 & 2.581 & 1.02 & 0.0138  \\
          & DSC+CA &  \textbf{27.54} & \textbf{0.919} & \textbf{0.592} & 2.595 & \textbf{0.86} & \textbf{0.0127} \\
    \bottomrule
  \end{tabular}
\end{table}
\subsubsection{Ablations about FAAM.} 
FAAM severs as the core module of UVENet. Therefore, we conduct detailed ablation studies about it from three aspects. Firstly, we gradually increase its usage to align and aggregate multi-frame feature maps at different scales, as shown in~\cref{tab:abla_faam}. When not applying FAAMs (first row in~\cref{tab:abla_faam}), the model performs the worst. Conversely, when FAAM is applied at all scales (last row in~\cref{tab:abla_faam}), the model performs the best. The observation indicates the importance of alignment at multiple scales, which can cover spatial shifts at different scales. Secondly, we adjust the key hyperparameter within FAAM, namely the base shift length $l$. An appropriate $l$ is also crucial for the spatial alignment of multiple frames. The first four rows in~\cref{tab:abla_faam1} show that the overall performance of the model peaks when $l=3$. Lastly, we investigate the effects of different aggregation methods on the model's performance. Depth-wise convolution is responsible for spatial aggregation, while point-wise convolution handles temporal aggregation. Results in~\cref{tab:abla_faam1} demonstrate removing either will degrade the performance. Moreover, channel attention further improves the model's performance.

\begin{figure}[t]
  \centering
  \includegraphics[width=0.99\columnwidth]{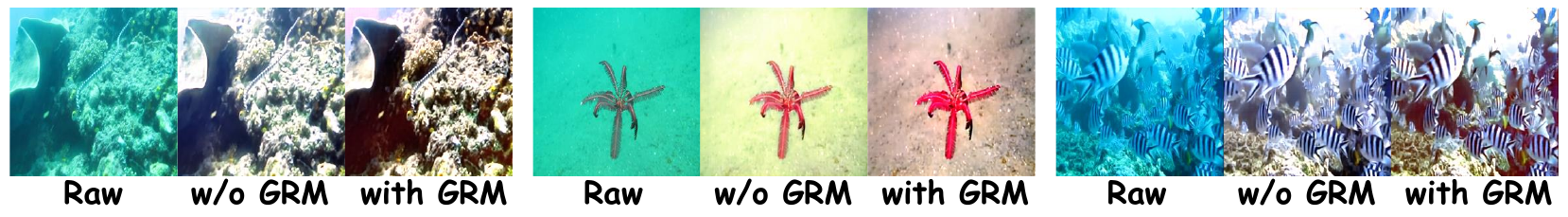}
  \caption{Visual comparison of ablation results on GRM.}
  \label{fig:abla_grm}
\end{figure}

\subsubsection{Effect of GRM.} 
As shown in~\cref{fig:abla_grm}, we visually compare the enhancement results of models with and without GRM. It can be observed that the frames enhanced by the model with GRM appear more natural in color, correcting some greenish and overly bright phenomena.

\section{Limitations and Future Work}
Firstly, there exists a domain gap between the synthetic underwater videos in the SUVE dataset and real underwater videos. The models trained on SUVE may not generalize well to real-world underwater scenarios, which is a common issue prevalent in UIE as well. Inspired by the solutions in UIE, domain adaptation techniques can help alleviate this issue~\cite{jiang2022two,wang2023domain}. Secondly, while our model UVENet is specifically tailored for underwater videos, its reliance on a large encoder (ConvNeXt) precludes its application in real-time video enhancement scenarios. Despite this efficiency limitation, it is important to acknowledge that UVENet represents a pioneering attempt in this field. Future efforts will focus on designing more lightweight models when maintaining the enhancement performance. Finally, we only consider single image quality metrics and temporal consistency metrics when evaluating. Designing more reasonable evaluation metrics for underwater videos is also a valuable research topic.

\section{Conclusion}
To address the challenge of applying single-image enhancement techniques to underwater videos, we systematically investigate the task of underwater video enhancement for the first time. To this end, we construct the Synthetic Underwater Video Enhancement (SUVE) dataset, comprising pairs of underwater-style videos and their corresponding high-quality reference videos, which can be utilized for supervised model training. Furthermore, we propose a novel underwater video enhancement model, UVENet, which leverages inter-frame relationships for enhancement. Experimental results demonstrate that UVENet trained end-to-end on the SUVE dataset can achieve superior enhancement performance for underwater videos. Our approach not only elevates the state-of-the-art but also paves the way for future advancements in underwater video enhancement.

%
%
\bibliographystyle{splncs04}
\bibliography{egbib}

\clearpage
\appendix
\section{More Details about UVENet}
In this section, we provide a more detailed description of the structure and hyperparameters of UVENet. The Encoder is a tiny version of ConvNeXt, where we only replaced Layer Normalization with Instance Normalization. The structure and hyperparameters of the encoder can be referred to in the original ConvNeXt paper. For the other parts of UVENet, the main learnable structures are the convolutional layers. \cref{tab:model} displays the hyperparameters of these convolutional layers. Here, T represents the number of input frames. The first two hyperparameters of each convolutional layer are the input and output channel numbers, while the subsequent three hyperparameters k, p, and g represent the kernel size, padding, and groups, respectively. For FAAM, the n\_feat is determined based on the scale it operates on. The n\_feat values for the four FAAMs are 96, 192, 384, and 768, respectively. Note that the two convolutional layers in DSC are depth-wise convolution and point-wise convolution. 

\begin{table}[ht]
  \caption{The detailed structures and hyperparameters of UVENet.}
  \label{tab:model}
  \centering
  \begin{tabular}{c|l}
    \toprule
    Component & Details \\
    \midrule
    \multirow{2}[1]{*}{DSC in FAAM} 
    & Conv (n\_feat*T, n\_feat*T, k=3, p=1, g=n\_feat*T) \\
    & Conv (n\_feat*T, n\_feat, k=1, p=0, g=1) \\
    \midrule
    \multirow{2}[1]{*}{CA in FAAM} 
    & Conv (n\_feat, n\_feat/16, k=1, p=0, g=1) \\
    & Conv (n\_feat/16, n\_feat, k=1, p=0, g=1) \\
    \midrule
    \multirow{7}[1]{*}{Decoder}  
    & Conv (768, 96, k=1, p=0, g=1)  \\
    & Conv (384, 96, k=1, p=0, g=1)  \\
    & Conv (192, 96, k=1, p=0, g=1)  \\
    & Conv (96, 96, k=1, p=0, g=1)  \\
    & Conv (384, 96, k=3, p=1, g=1)  \\
    & Conv (96, 96, k=3, p=1, g=1)  \\
    & Conv (96, 3, k=3, p=1, g=1)  \\
    \midrule
    \multirow{4}[1]{*}{GRM} 
    & Conv (3*T+3, 64, k=3, p=1, g=1)  \\
    & Conv (64, 64, k=3, p=1, g=1)  \\
    & Conv (64, 64, k=3, p=1, g=1)  \\
    & Conv (64, 3, k=3, p=1, g=1)  \\
  \bottomrule
  \end{tabular}
\end{table}

\section{More Experiments}

\begin{figure}[t]
  \centering
  \includegraphics[width=0.9\columnwidth]{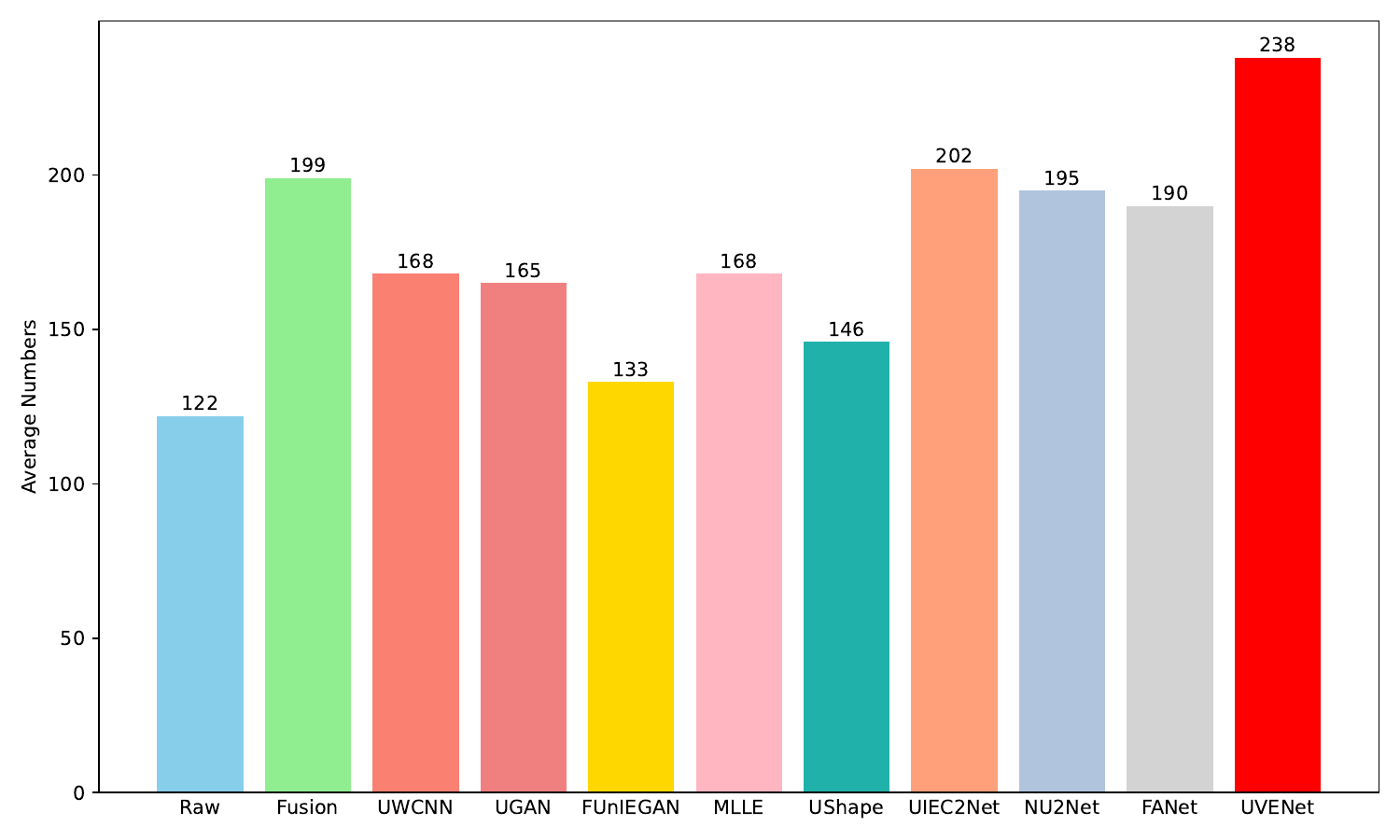}
  \caption{Average number of matched key points for various methods.}
  \label{fig:sift}
\end{figure}

\begin{figure}[t]
  \centering
  \includegraphics[width=0.95\columnwidth]{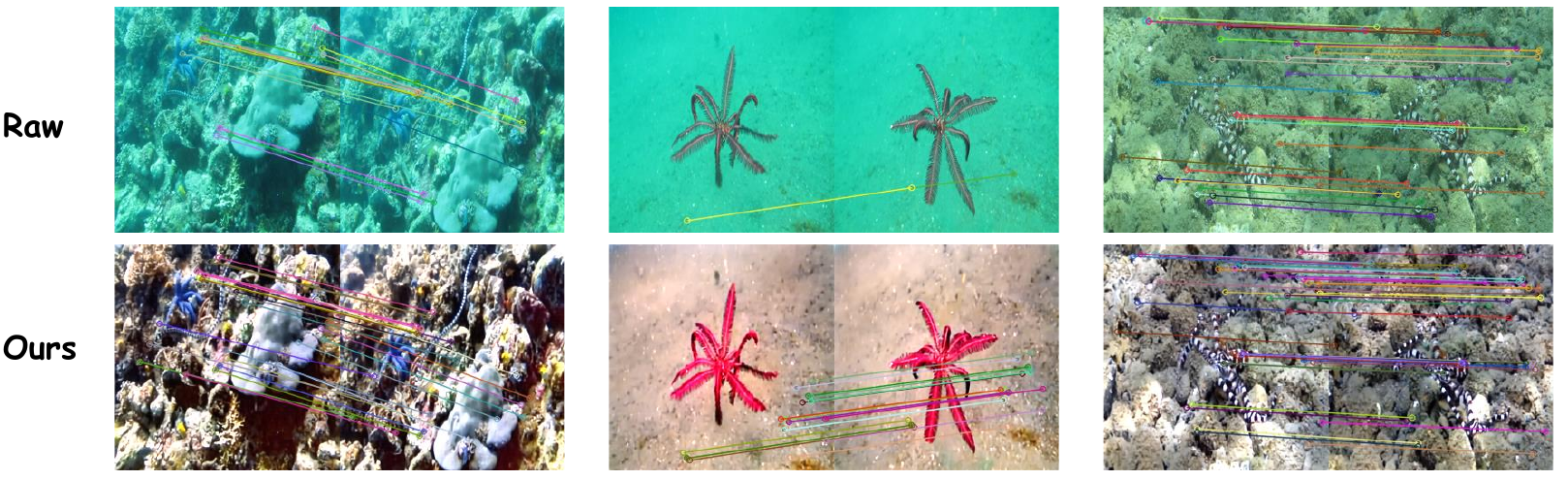}
  \caption{Comparison of SIFT key points matching.}
  \label{fig:keypoint}
\end{figure}

In this section, we focus on the representation performance of various UIE algorithms and our UVENet. We randomly sample many pairs of temporally close frames from the real underwater videos in the MVK dataset. To increase the matching difficulty, the interval between each pair of frames is 6 frames apart rather than being completely adjacent frames. Subsequently, we extract key points using the scale-invariant feature transform (SIFT) algorithm and match key points for each pair of frames. \cref{fig:sift} illustrates the average number of matched key points for various methods. The baseline (RAW) performs the worst, while our UVENet performs the best. This experiment demonstrates that underwater videos enhanced by UVENet not only exhibit superior visual effects but also prove beneficial for downstream tasks. \cref{fig:keypoint} presents three visual examples of the SIFT matching on raw image pair and enhanced results, where we filtered the matched points using the same distance threshold. We can also infer the direction of camera motion from the relative positions of matched key points.

\section{More Samples in SUVE}
\begin{figure}[t]
  \centering
  \includegraphics[width=0.9\columnwidth]{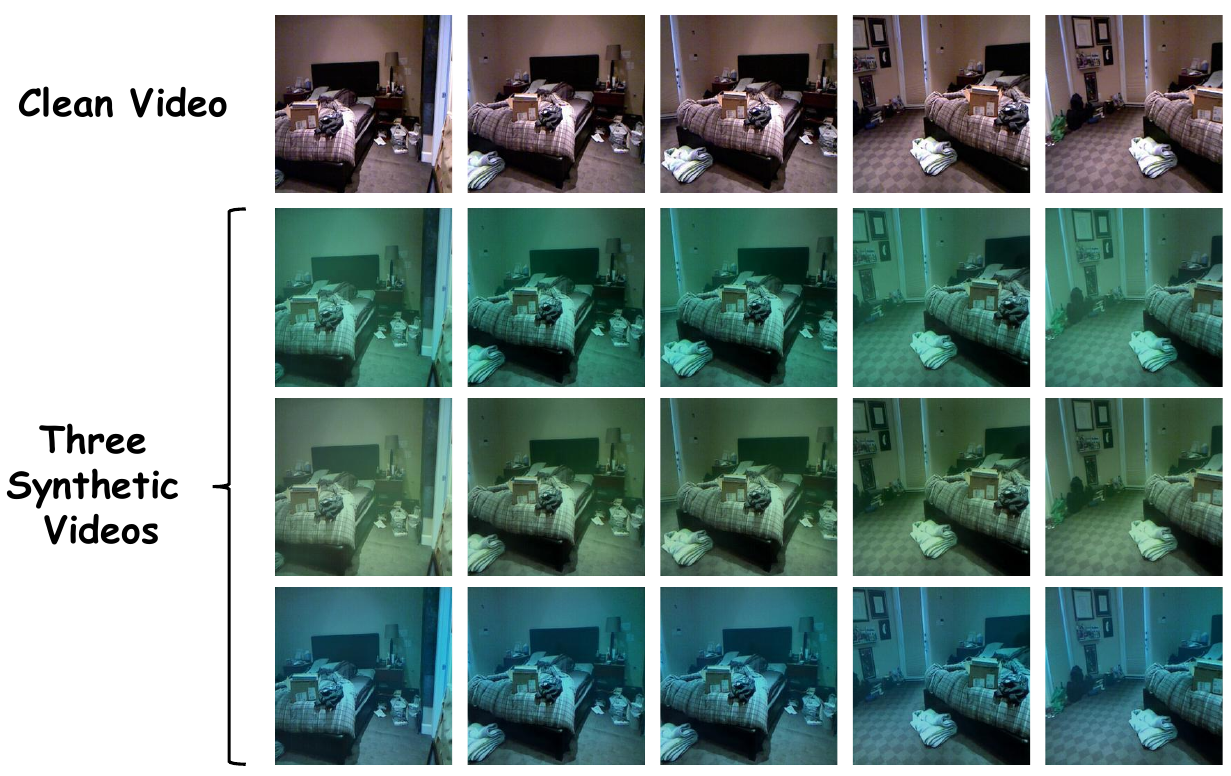}
  \caption{One clean video and its corresponding three types of synthetic underwater-style videos in the SUVE dataset. We randomly select five frames for ease of display.}
  \label{fig:suve1}
\end{figure}

\begin{figure}[t]
  \centering
  \includegraphics[width=0.9\columnwidth]{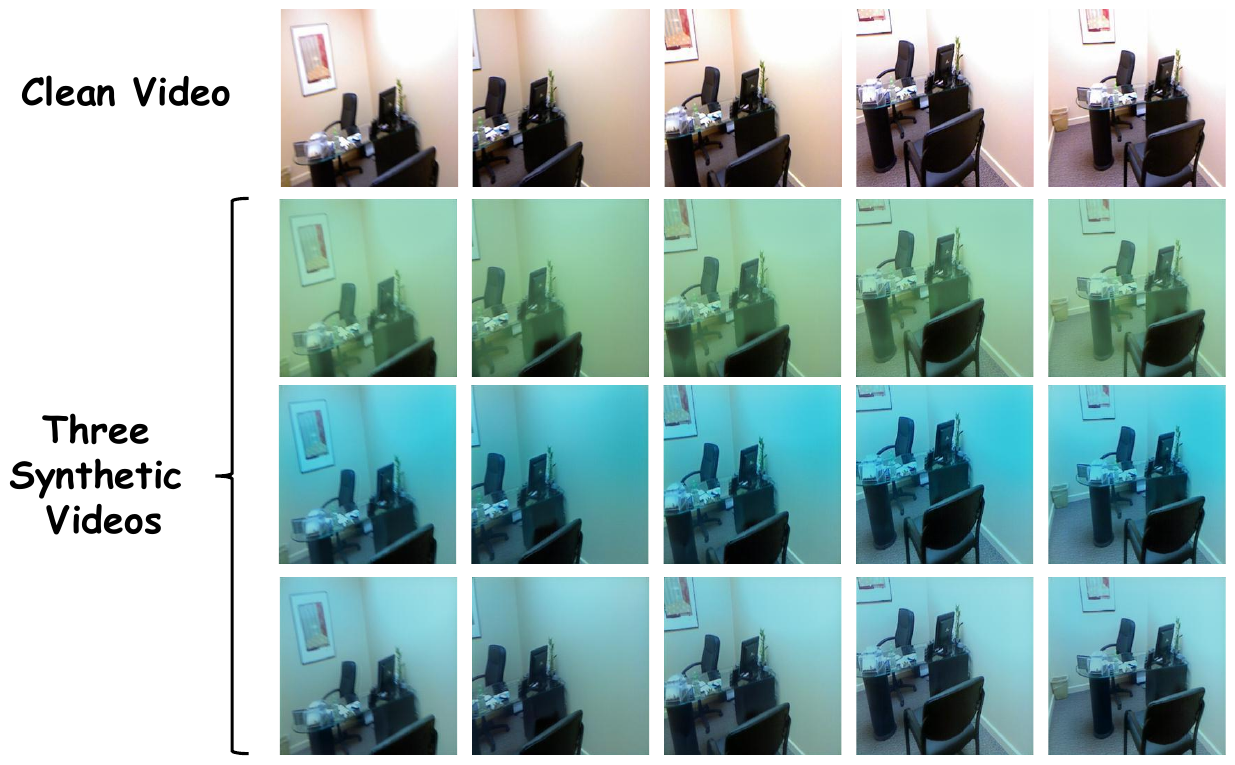}
  \caption{One clean video and its corresponding three types of synthetic underwater-style videos in the SUVE dataset. We randomly select five frames for ease of display.}
  \label{fig:suve2}
\end{figure}

We visualize more synthetic video samples in~\cref{fig:suve1,fig:suve2}.

\section{More Visual Results}

We provide more visual results in~\cref{fig:exp_mvk1,fig:exp_mvk2}. Four demo videos are also provided in the supplementary materials, visualizing the enhancement results of video samples from the SUVE and MVK datasets.

\begin{figure}[t]
  \centering
  \includegraphics[width=0.99\columnwidth]{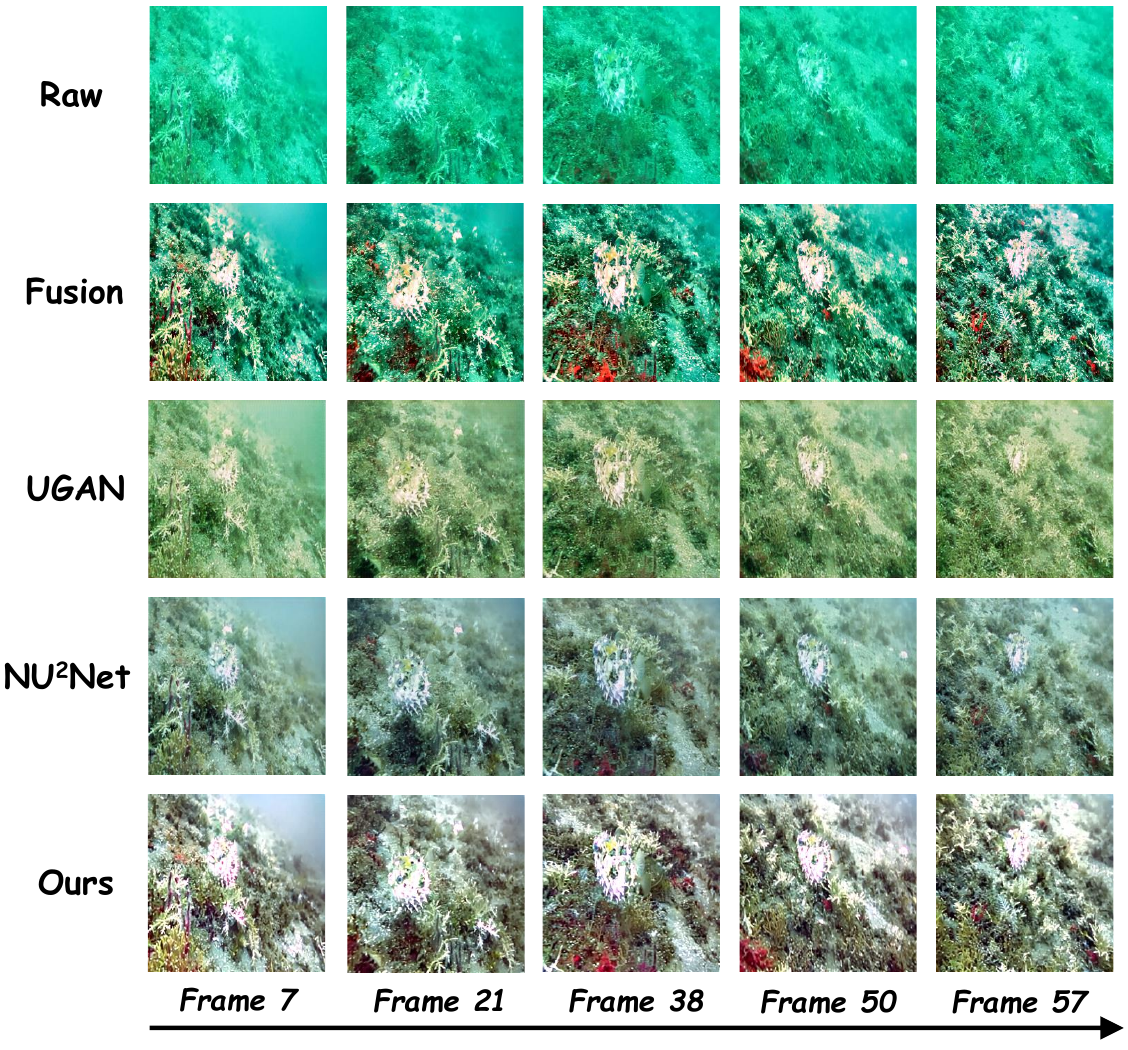}
  \caption{Visual comparison of our UVENet with several UIE methods on one real underwater video from the MVK dataset. We randomly select five frames for ease of display.}
  \label{fig:exp_mvk1}
\end{figure}

\begin{figure}[t]
  \centering
  \includegraphics[width=0.99\columnwidth]{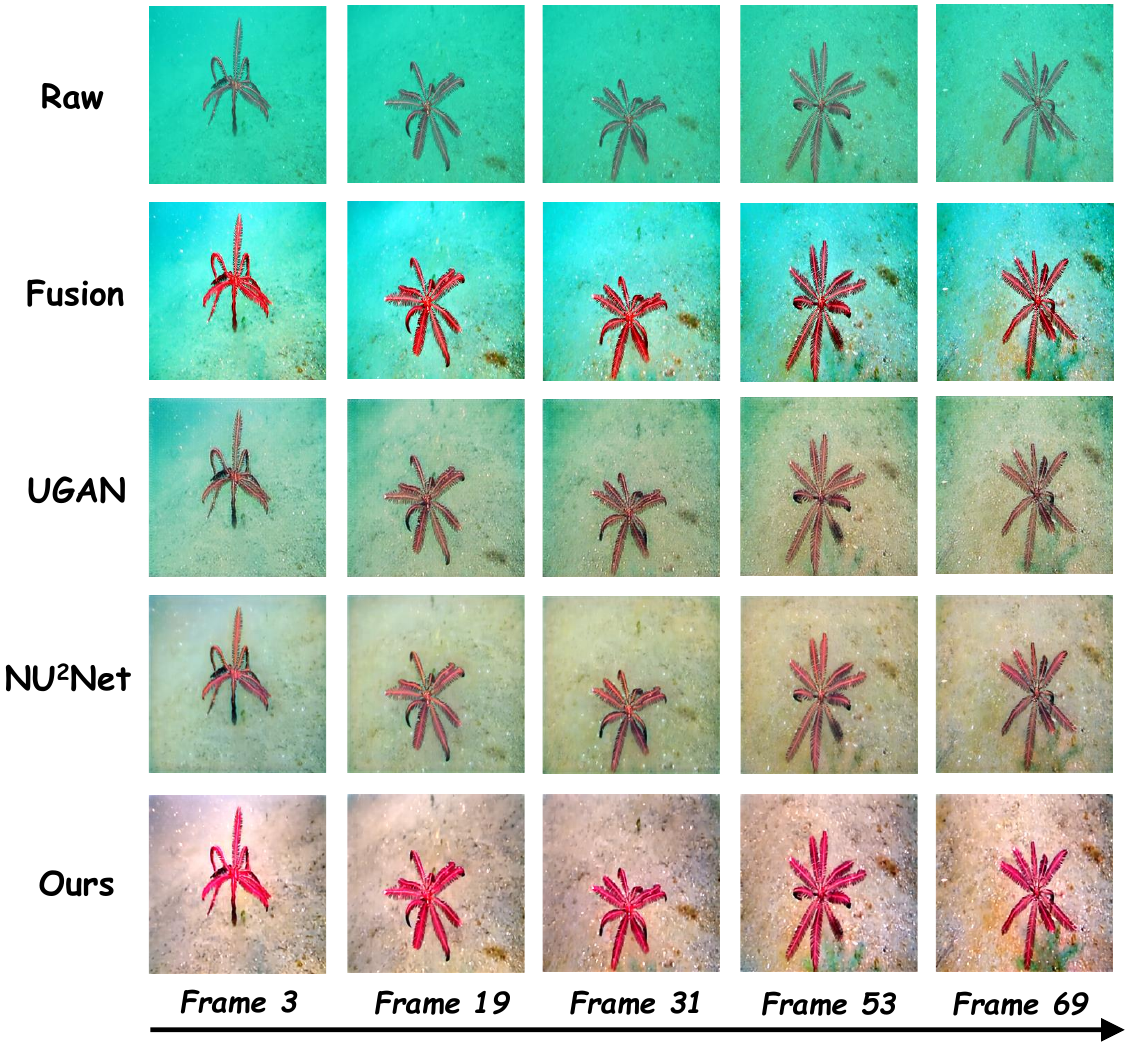}
  \caption{Visual comparison of our UVENet with several UIE methods on one real underwater video from the MVK dataset. We randomly select five frames for ease of display.}
  \label{fig:exp_mvk2}
\end{figure}

\end{document}